\documentclass{article}

\PassOptionsToPackage{numbers, compress}{natbib}

\usepackage[final]{nips_2018}

\usepackage[utf8]{inputenc} 
\usepackage[T1]{fontenc}    
\usepackage{hyperref}       
\usepackage{url}            
\usepackage{booktabs}       
\usepackage{amsfonts}       
\usepackage{nicefrac}       
\usepackage{microtype}      
\usepackage{amsmath}
\usepackage{graphicx}
\usepackage{subcaption}
\usepackage{textcomp}
\usepackage[export]{adjustbox}

\title{Relation Networks for Optic Disc and Fovea Localization in Retinal Images}

\author{
    Sudharshan Chandra Babu\footnotemark[1]\\
    Department of Computer Science\\
    Indian Institute of Technology Bombay\\
    Mumbai, Maharashtra 400076 \\
\texttt{cbsudu@gmail.com} \\
   \And
    Shishira R Maiya\thanks{Joint First authors.} \\
    Robert Bosch Center for Cyber Physical Systems \\
    Indian Institute of Science\\
    Bangalore, Karnataka 560054 \\
   \texttt{shishirar@iisc.ac.in} \\
   \And
    Sivasankar Elango \\
    Department of Computer Science and Engineering\\
    National Institute of Technology, Tiruchirappalli\\
    Tiruchirappalli, Tamil Nadu 620015 \\
  \texttt{sivasankar@nitt.edu} \\
}

\begin{document}

\maketitle

\begin{abstract}
Diabetic Retinopathy is the leading cause of blindness in the world. At least 90\% of new cases can be reduced with proper treatment and monitoring of the eyes. However, scanning the entire population of patients is a difficult endeavor. Computer-aided diagnosis tools in retinal image analysis can make the process scalable and efficient. In this work, we focus on the problem of localizing the centers of the Optic disc and Fovea, a task crucial to the analysis of retinal scans. Current systems recognize the Optic disc and Fovea individually, without exploiting their relations during learning. We propose a novel approach to localizing the centers of the Optic disc and Fovea by simultaneously processing them and modelling their relative geometry and appearance. We show that our approach improves localization and recognition by incorporating object-object relations efficiently, and achieves highly competitive results.

\end{abstract}

\section{Introduction}
Diabetes mellitus or diabetes is a group of disorders caused by very high blood sugar levels over a long period of time. Diabetic retinopathy is the leading cause of blindness in the world. At least 90\% of new cases could be reduced with proper treatment and monitoring of the eyes \citep{Tapp2003ThePO}. The resulting blindness can often be prevented by timely and periodic checking of the retinal scans for any abnormalities in the optic disc and fovea. Their identification and localization is an important requisite for the segmentation of blood vessels and in the detection of other abnormalities \cite{kaur2012automated}.  However, scanning the entire population of patients is a difficult endeavor. Computer-aided diagnosis tools in retinal image analysis can make the process scalable and speed things up considerably.  

It can be observed that in most fundus images, the optic disc and fovea have a constant relative geometry, i.e their relative spatial positions and shape are mostly constant. Traditionally, ophthalmologists have used the relative geometry between the optic disc and fovea for determining the location of the fovea. Current systems recognize the optic disc and fovea individually \citep{raja2014automatic, akram2010retinal, alghamdi2016automatic}, and do not exploit their relations during learning. 

Motivated by the limitations of current work, we incorporate an attention based object relation module \citep{hu2017relation}, and propose a novel multi-stage framework for the localization of optic disc and fovea centers by learning the intrinsic object-object relationships between the two entities. We break down the problem into two stages--local object detection, followed by point regression, i.e, we first detect the optic disc and fovea, following which we regress to their centers. In the first stage, the relation module enables the optic disc and fovea to be processed and reasoned simultaneously, instead of being recognized individually and this joint reasoning improves the recognition accuracy. The second stage consists of a simple regressor, that regresses the joint coordinates from the bounding box regions predicted in the first stage. We conduct our study on the IDRiD dataset \citep{porwal2018indian} and demonstrate highly competitive results on the test set and the IDRiD competition leaderboard \citep{leaderboard}.

Our framework need not be limited to optic disc and fovea detection. The idea of performing joint reasoning and utilising object-object relations to improve recognition is highly relevant in the field of medical imaging, where organs, tissues etc are geometrically correlated and have a learnable object-object relationship.

\section{Related work}

Early work on optic disc and fovea center localization focused on using traditional methods such as, thresholding and centroid calculation of connected components \cite{raja2014automatic}. Akram et al. \citep{akram2010retinal} preprocess the green channel of the image and use a histogram based method to localize. Raja et al. \cite{raja2014automatic}  enhance contrast using CLAHE \cite{pizer1987adaptive} and perform localization. These methods have been tested on small datasets like \cite{staal2004ridge, hoover2000locating, kamarainen2007diaretdb1} and are not robust.

Deep Convolutional Networks have also been used to detect the optic disc and fovea. \cite{engr2018OpticDD, Sevastopolsky2017OpticDA, alghamdi2016automatic}. Alghamdi et al. \citep{alghamdi2016automatic} propose a simple 5 layer CNN, along with a cascade feature classifier for identifying the optic disc and fovea. However, these methods detect the optic disc and fovea independently and do not utilize the relationship between them. Our proposed approach learns the relationship between the optic disc and fovea, and improves upon the existing approaches. Our approach enables joint reasoning of the optic disc and fovea and considerably improves recognition accuracy.

\begin{figure}
  \includegraphics[width=0.99\linewidth]{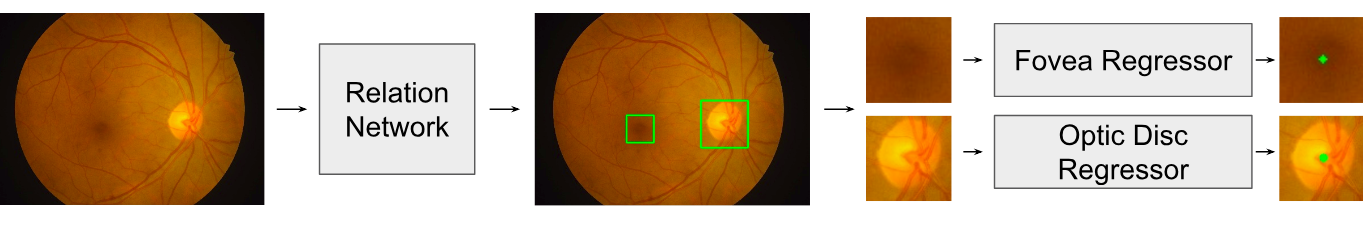}
  \caption{A schematic illustration of our multi-stage method: The input retinal image is passed through the Relation Net. Crops are extracted and regressors are trained to localize the centers.}
  \label{fig:net}
\end{figure}

\section{Approach}
\subsection{Dataset and Pre-processing}
 
We use the Indian Diabetic Retinopathy Image Dataset (IDRiD) \citep{porwal2018indian} for our experiments, which consists of 516 images--413 for training and 103 for testing. All images have a paired list of ground truth center coordinates. The original images have a resolution of 4288x2848.  We downsample these images to 512x512 and normalize the training data. We also apply Contrast Stretched Adaptive Histogram Equalization (CLAHE) \cite{pizer1987adaptive} to enhance the contrast of each image. We perform data augmentation by translation, shearing, scaling and horizontal flipping. To train the first stage, i.e the Relation Network, we generate ground truth bounding boxes around each optic disc and fovea instance, using the ground center coordinates. We thus prepare a dataset of ground truth images along with their corresponding centers and bounding box annotations. 


\subsection{Optic Disc and Fovea Localization}

Our approach consists of two stages. The first stage of our pipeline is an object detection task--to detect the the optic disc and fovea. The output of the first stage are bounding boxes predictions. Using the predicted bounding boxes, we extract image crops and train a simple regressor to regress to the coordinates, as illustrated in Figure \ref{fig:net}. The first stage of our pipeline is the Relation Network for Object detection \citep{hu2017relation}. We compare our approach against two baselines. We use a ResNet-18 and a ResNet-50  to regress directly on the point coordinates. We apply the same pre-processing and data augmentation to the baselines. 

\subsubsection{Relation Network}

The Relation Network uses a Faster-RCNN \cite{ren2015faster} with Resnet-101 \cite{he2015deep} as the feature extractor. The Relation Network consists of object relation modules, which is an attention based fully differentiable module. The proposed module extends the original attention weight into two components: the original weight and a new geometric weight. The latter models the spatial relationships between objects and only considers the relative geometry between them, making the module translation and rotation invariant. The relation module enables joint reasoning of all objects and improves recognition accuracy.

\begin{align}
\label{eq:1}
f_{R}(n) = \sum_{m}  \omega^{mn} \cdot (W_{v} \cdot f_{A}^m)
\end{align}

Eq. (\ref{eq:1}) shows how the relation module computes a feature vector of all objects in the image with respect to the \begin{math}
n_{th}    \end{math} object in the image. \begin{math} W_{v} \end{math}
linearly transforms the sum of appearance feature vectors \begin{math}
f_{A}^m\end{math}. The relation weight \begin{math} \omega^{mn} \end{math} indicates the impact of other objects on the \begin{math} n_{th} \end{math} object. Eq. (\ref{eq:1}) shows the output of one relation module.
The relation vectors from many such modules are calculated for all objects in the image and is concatenated with the input appearance feature vector as shown in Eq. (\ref{eq:2}).

\begin{align}
\label{eq:2}
f_{a}^n = f_{a}^n + concat[f_{r}^{1}(n),....,f_{r}^{N_{r}}(n)]
\end{align}
\bigbreak 

The output dimensions of the feature appearance vector \begin{math} f_{a}^n \end{math} is unchanged after these operations with the relation module and we proceed to use fully connected layers on them to predict bounding box and confidence scores. Also, the traditional NMS method for duplicate removal is replaced and improved by a lightweight relation module.

\subsubsection{Simple Regressor}

The second stage of our approach consists of extracting crops from the image, based on the predicted bounding boxes from the previous stage. These crops are fed to a simple regressor--A two layer CNN with batch normalization \cite{Ioffe2015BatchNA}. We train this network to regress the coordinates of the optic disc and fovea centers.

\subsection{Training Details}

We used an open source implementation of Relation Network \cite{hu2017relation} and implemented the regressors in PyTorch \citep{paszke2017automatic}. We trained the Relation Network for 8 epochs. We use SGD as the optimizer for the Relation Network, with a weight decay of $10^{-4}$
and momentum of 0.9. We train the regressors using Adam. We trained the models on a Nvidia 1080 Ti GPU.

\section{Results}

\begin{figure}[h!]
  \centering
  \begin{subfigure}[b]{0.32\linewidth}
    \includegraphics[width=\linewidth]{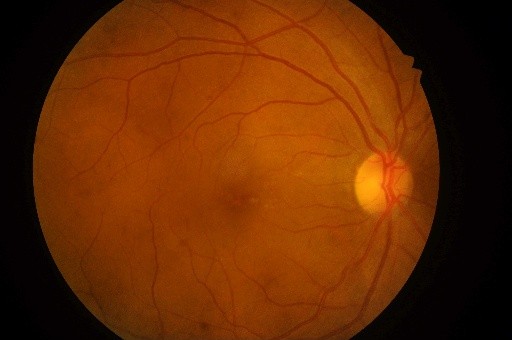}
  \end{subfigure}
    \begin{subfigure}[b]{0.32\linewidth}
    \includegraphics[width=\linewidth]{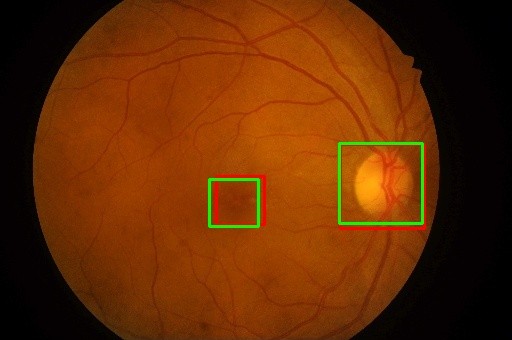}
  \end{subfigure}
    \begin{subfigure}[b]{0.32\linewidth}
    \includegraphics[width=\linewidth]{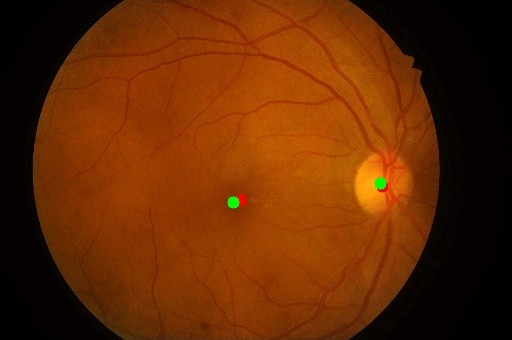}
  \end{subfigure}
    \begin{subfigure}[b]{0.32\linewidth}
    \includegraphics[width=\linewidth]{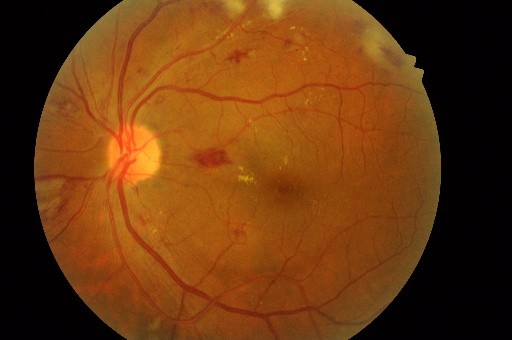}
  \end{subfigure}
      \begin{subfigure}[b]{0.32\linewidth}
    \includegraphics[width=\linewidth]{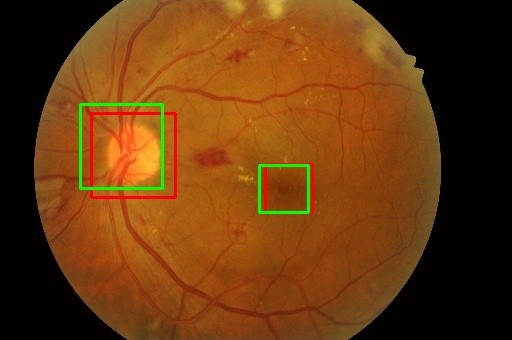}
  \end{subfigure}
    \begin{subfigure}[b]{0.32\linewidth}
    \includegraphics[width=\linewidth]{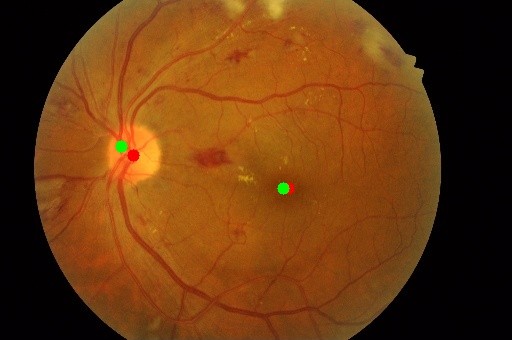}
  \end{subfigure}
  \caption{Qualitative results on test set. Predictions in \textbf{green} and ground truth annotations are in \textbf{red}. Input image (left), bounding boxes (middle) and localized center coordinates (right).}
  \label{fig:gt_preds}
\end{figure}

For evaluation, we use Mean Average Precision (mAP) to measure the accuracy of object detection and mean Euclidean Distance between the predicted coordinates and the ground truth coordinates to measure the accuracy of localization. We demonstrate strong results on these metrics on the test set as seen in Table 1, Table 2 and Figure \ref{fig:gt_preds}. 

The testing set consists of 103 images of resolution of 4288x2848 with ground truth coordinates. We apply the same resizing and pre-processing that we applied to the training set during testing. Following our training pipeline, we pass the testing images through the Relation Network and obtain bounding boxes. We extract image crops and regress on them to get optic disc, and fovea coordinate predictions. Once we have the predictions, we rescale them to the original pixel coordinate space (4288x2848) and evaluate against the ground truth. 

\begin{table}[h!]
 \caption{Evaluation of object detection by Relation Network. Results are reported in Mean Average Precision (mAP)}

 \begin{center}
 \begin{tabular}{l l}
 \toprule
 \multicolumn{2}{c}{\textbf{Relation Network}} \\ 
     \midrule
        mAP$_{\text{50:95}}$ & 69.4 \\
        mAP$_{\text{50}}$ & 97.3 \\
        mAP$_{\text{75}}$ & 80.2 \\
      \bottomrule
\end{tabular}
\end{center}
\end{table}

We compare our approach against the two baselines. We call our combined approach Relation Network Regressor and demonstrate superior results over the baselines as seen in Table 2. We also achieve highly competitive results on the IDRiD Localization Challenge.  \citep{leaderboard}.



\begin{table}[h!]
 \caption{Localization results on the test dataset. The numbers are mean Euclidean Distance between ground truth and predicted coordinates.}

 \begin{center}
 \begin{tabular}{l l l l l}
 \toprule
\textbf{Model} &\textbf{Optic Disc} &\textbf{Fovea}\\
     \midrule
      ResNet-18  & 80.48 & 115.22\\
      ResNet-50  & 60.32 & 95.45\\
      1st place approach & \textbf{25.61}  & 45.89 \\
      Relation Network Regressor  & 26.12 & \textbf{43.46}\\
      \bottomrule
\end{tabular}
\end{center}
\end{table}

\section{Conclusion}

In this work, we present a novel approach for the detection of optic disc and fovea, and the localization of their centers using a Relation Network. We show that our approach learns the relationship between the optic disc and fovea and applies joint-reasoning to improve recognition. We show that our model demonstrates strong qualitative results and achieves a highly competitive performance in this task.

\bibliography{references}
\bibliographystyle{ieeetr}

\end{document}